\documentclass[journal]{IEEEtran}
\usepackage{blindtext}
\usepackage{graphicx}
\usepackage{float}
\usepackage[justification=centering]{caption}
\graphicspath{ {/home/gopeshh/Pictures/} }
\ifCLASSINFOpdf
\else
\fi
\usepackage{amsmath}

\DeclareMathOperator*{\argmin}{arg\,min}

\usepackage{physics}

\usepackage[ruled, linesnumbered]{algorithm2e}
\usepackage[justification=centering]{caption}
\usepackage{subfigure} 
\usepackage{algorithmic}

\begin{document}
%
\title{Trajectory Control for Differential Drive Mobile Manipulators}
%
%
%

\author{\normalsize Harish~Karunakaran,
        Gopeshh~Raaj~Subbaraj,
        Robotics~Engineering,~WPI
        }

\maketitle

\begin{abstract}
Mobile manipulator systems are comprised of a mobile platform with one or more manipulators and are of great interest in a number of applications such as indoor warehouses, mining, construction, forestry etc. We present an approach for computing actuator commands for such systems so that they can follow desired end-effector and platform trajectories without the violation of the nonholonomic constraints of the system in an indoor warehouse environment. We work with the Fetch robot which consists of a 7-DOF manipulator with a differential drive mobile base to validate our method. The major contributions of our project are, writing the dynamics of the system, Trajectory planning for the manipulator and the mobile base, state machine for the pick and place task and the inverse kinematics of the manipulator. Our results indicate that we are able to successfully implement trajectory control on the mobile base and the manipulator of the Fetch robot. 
\end{abstract}


%
\section{Introduction}
A mobile manipulator is defined as a robotic system composed of a mobile platform and a manipulator mounted on the platform equipped with non-deformable wheels. This combined system of a manipulator and mobile base is able to perform manipulation tasks in a much larger workspace than a fixed-base manipulator such as the Kuka Manipulator. Conventional robot manipulators with a fixed base have limited reachable workspace. Thus the addition a wheeled mobile platform to a robot manipulator provides added mobility to the robotic system. Typically, a mobile manipulator consists of a multi-link manipulator and a mobile platform. This increases the effective workspace of the end-effector of the manipulator when compared to fixed base systems. In recent years, these mobile manipulators have become a subject of interest because of the mobility and dexterity. Thus, many service robots employ the similar design which we use in our project. 

The integration of a manipulator and a mobile platform, however, gives rise to difficulties such as decomposing a given task into fine motions to be carried out by the manipulator and the gross motions to be achieved by the mobile platform and how to establish the dynamic model of the mobile manipulator system in a systematic approach. Since, in most cases, the mobile platform is subjected to nonholonomic constraints and the robot manipulator is a holonomic mechanical system, how to control the manipulator and the mobile platform to avoid possible tip over is also important [10]. A nonholonomic system is a system whose state depends on the path taken in order to achieve it. Also, unlike conventional industrial manipulators mounted on stationary bases, a mobile manipulators motions interact dynamically with its base robot, which degrades system performance, resulting in performance problems like excessive end-effector errors and poor system stability. These systems operate in highly unstructured environments, thus limiting the various sensing techniques available for their control. The above characteristics of mobile manipulators present challenging control problems.
 
A Differential Drive robot is a mobile robot whose movement is based on two separately driven wheels placed on either side of the robot body. It can thus change its direction by varying the relative rate of rotation of its wheels and hence does not require an additional steering motion. A differential drive robot
cannot move in the direction along the axis which is a singularity. They are also very sensitive to slight changes in velocity in each of the wheels. Even small changes in the relative velocities between the wheels affects the robot's trajectory. This makes the control of differential drive system difficult to implement in practice.

Our project focuses on trajectory following and control for mobile manipulators and tries to address some of the problems discussed above. We develop a control method for manipulators mounted on mobile robots which applies to large manipulator motions when system characteristics are highly variable and when the accuracy and the availability of sensory information is limited. The Fetch mobile manipulator platform is used in a simulated environments which is a differentially driven platform, equipped with a 7-DOF manipulator is used. The Fetch robot is a combination of two subsystems. The torso of this robot, called the Fetch, comprises of a camera and other vision sensors along with the 7-DOF manipulator, while the base, called the Freight, enables the navigation of the entire robot. The torso is mounted on the freight. Fetch Robotics’ system uses a relatively large and capable mobile manipulator to pick items off of warehouse shelves, while Freight, acts as an autonomous cargo delivery cart. Fetch can pick items continuously, while a succession of Freights can switch in and out to move different selections of goods to different parts of the distribution centre. The Fetch robot is shown in Figure 1.

The Denavit-Hartenburg parameters for the 7-DOF manipulator is found out and is incorporated as part of the URDF (Unified Robot Description) of the fetch robot. Lagrange’s method is used to obtain the reduced equations of motion from the base of the robot to the end-effector. This gives us the dynamics of the system. Next, we perform trajectory planning for the arm of the fetch robot to plan from one joint position to another. Trac\_ik[11] package available as an open source implementation is used to find the inverse kinematics of the arm. Then, for moving the mobile base of the fetch robot, we used a planning algorithm to find a path from a start position in the simulated environment to a goal position. In order to make the mobile base and the trajectory planning work in conjunction, we developed a state-machine for a pick and place task. The performance of our method has been tested on this task in a simulated environment in Gazebo and has been discussed in the later sections.

\begin{figure}[h]
\centering
\includegraphics[width=0.4\textwidth]{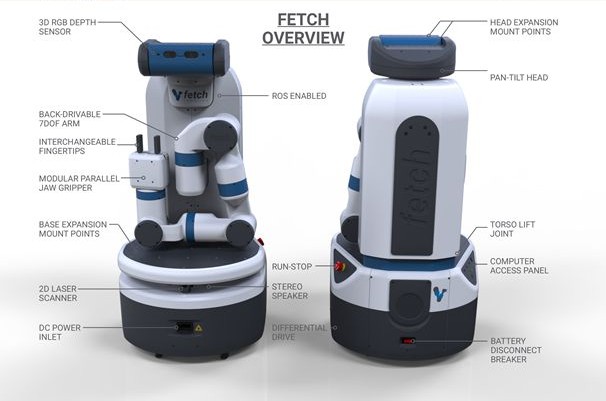}
\caption{Fetch Robot}
\end{figure}

The applications of our system extend into the warehouse domain where trajectory planning and control is an important aspect of picking and placing objects in a warehouse or an industrial setting. This mobile manipulator control method would provide greater autonomy to service robots and extend their use. Our method tries to solve the problem of control in mobile manipulators described above using the Fetch robot as an example. The rest of this paper is divided as follows, the related work is discussed in Section II followed by the Methodology in Section III and then the results and discussion are presented in Section IV followed by Conclusion and Future work in Section V. 

\section{Related Work}

In the past issues related to Mobile manipulators such as dynamic and static stability, force development and application control in the presence of base compliance, dynamic coupling issues, etc. have been studied [2]. Alicja et al.[4] worked on the problem of trajectory tracking of non-holonomic mobile manipulators. They presented two versions of the tracking control algorithm, one for mobile manipulators with a non-holonomic platform only and the other for a doubly non-holonomic mobile manipulator. They then compared the control process of these in a simulation environment and made some suggestions about a choice of drives for the robotic arm.  They presented a dynamic control algorithm for holonomic systems in general which can be used for a mobile platform with any robot mounted atop of it. This was one of the earlier approaches to tackle the problem of trajectory tracking as the computing power was limited back then.

R.Solea et al.[5] in their paper focus on the motion planning problem of mobile manipulator systems where they propose a methodology for generating trajectories for both the mobile platform and the manipulator that will take a system from an initial configuration to a pre-specified final one, without violating the non-holonomic constraint.  In order to ensure the smooth movement of the considered system they used nonlinear sliding mode control to solve the motion planning problem. They also present a controller design for the trajectory-tracking problem using the sliding mode control for a mobile platform equipped with a manipulator.

Real-time trajectory planning using model predictive control with constraints was proposed by Ide.S et. al. [6] for mobile manipulators. Their method uses Quadratic Programming (QP) for optimizing control inputs. The control inputs and outputs are limited corresponding to the required motion and the hardware specifications of the mobile manipulator. The required motion is changed frequently according to the situation and the hardware limitations, concretely the torque, the angular velocity, the mobile base velocity and the acceleration, which are subject to the hardware design. Their method was implemented to a real mobile manipulator model and real-time trajectory modification was demonstrated on a real mobile robot. But one of the drawbacks in their approach is that they generalize all manipulators and do not consider system specific parameters. But, in this project we use system parameters that are specific to the fetch robot as described in their approach.

Dong, W. et al [7] studied the tracking control problem of mobile manipulators considering the interaction between the mobile platform and the manipulator. They proposed a global tracking controller based on the dynamics of the defined tracking error and the extended Barbalat’s lemma. Their proposed controller ensures that the full state of the system asymptotically track the given desired trajectory globally in the presence of the system coupling.

Korayem, M.H. et al [8] in their paper proposed an Open-loop optimal control method as an approach for trajectory optimization of flexible mobile manipulator for a given two end-point task in point-to-point motion. Dynamic equations were derived using combined Euler Lagrange formulation which is similar to the approach we follow in our project. To solve the optimal control problem, they adopt an indirect method via establishing the Hamiltonian function and deriving the optimality condition from Pontryagins minimum principle. The equations obtained provided a two point boundary value problem which they then solved by numerical techniques.

Korayem, M.H.,et al [9] in their paper presented a methodology for collision-free trajectory planning of wheeled mobile manipulators in obstructed environments by means of potential functions. But in our case we do not consider obstructed environments making our approach simpler than theirs. In their method, all mobile manipulator parts and environmental obstacles were modeled as ellipsoids. Due to collision avoidance, the ellipsoid equations were expressed in a reference coordinate system and the corresponding dimensionless potential functions were
defined. Then, the trajectory planning of a spatial mobile robot in cluttered environment was performed, employing optimal control theory. One of the problems with their approach was that since they use control theory for trajectory planning, the planning times are higher than standard methods.

Inverse kinematics is an extensive field of research where numerous sophisticated algorithms have been presented during the last  decades.  However,  no  universal solution  could yet  be  found.  Most  prominent  approaches  either  attempt to  directly  derive  a  geometric  analytical  solution  from  the kinematic model or aim to solve the problem iteratively by optimization. Some of the most widely known and popular methods are based on computing the Jacobian using the transpose, pseudoinverse or damped least squares method [12]. The solution is  then  found  by following the gradient, but problems are often encountered by getting stuck in suboptimal extrema depending on the geometric complexity of the kinematic model. Constructive approaches successfully  overcome this issue by performing heuristic restarts from random initial configurations where [11] recently  proposed a novel method, TRAC-IK [11] that is publicly available under the ROS framework[13]. In this, significant performance increases regarding the success rate could be obtained over the classical Orocos KDL solver[15]. Their Trac\_ik method improves upon several failure points of KDL’s implementation of joint-limited pseudoinverse Jacobian inverse kinematics. They are able to outperform the KDL IK implementation by adding local minima and mitigation by reformulating the Inverse Kinematics problem as a Sequential Quadratic Programming problem and utilizing Cartesian tolerances to speed up the IK search. They are able to report a success rate of around 99.5\% on the Fetch robot and therefore is the choice for our IK solver.

\section{Methodology}

The overall work can be divided into the following modules,
\begin{itemize}
  \item Writing the Dynamics of the system
  \item Trajectory Planning of the mobile base and the manipulator
  \item Inverse Kinematics of the manipulator
  \item Finite State Machine for the pick and place task
  \item Trajectory Control using PID
\end{itemize}

\subsection{System Dynamics}
\begin{figure}[h]
\centering
\includegraphics[width=0.45\textwidth]{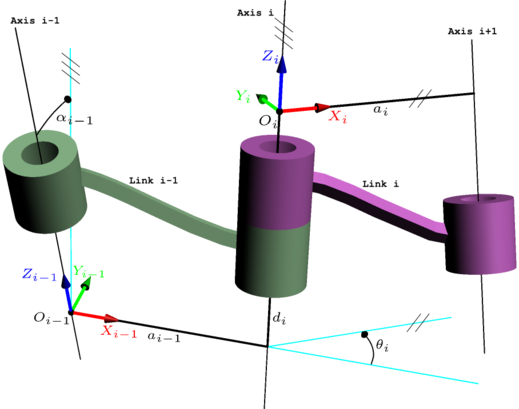}
\caption{The four parameters according to the DH convention are shown in red, $\theta_{i} , d_{i} , a_{i} , \alpha_{i}$}
\end{figure}

The first step in the writing the system dynamics involved writing the Denavit-Hartenberg (DH) parameters for the manipulator arm. DH parameters are the four parameters associated with a particular convention for attaching reference frames to the links of a spatial kinematic chain, or robot manipulator. This convention is a standard method for selecting frames of reference in robotics applications. This method is very useful to carry out kinematics operations in a manipulator. The four parameters of DH convention are used for transformation from one frame to another. 

These four parameters are :
\begin{itemize}
  \item d\,: offset along previous z to the common normal 
  
  \item $\theta$ \,: angle about previous  z, from old  x to new  x
  
  \item r\,: length of the common normal 
  \item$\alpha$ \,: angle about common normal.

\end{itemize}

The four parameters that associates two consecutive frames are shown in the figure 2. In this project we calculated the DH parameters for the fetch robot as the frames defined in the URDF file of fetch robot was not consistent with the DH convention. Figure 3 shows the frames for our system defined as per DH convention.

\begin{figure}[h]
\centering
\includegraphics[width=0.45\textwidth]{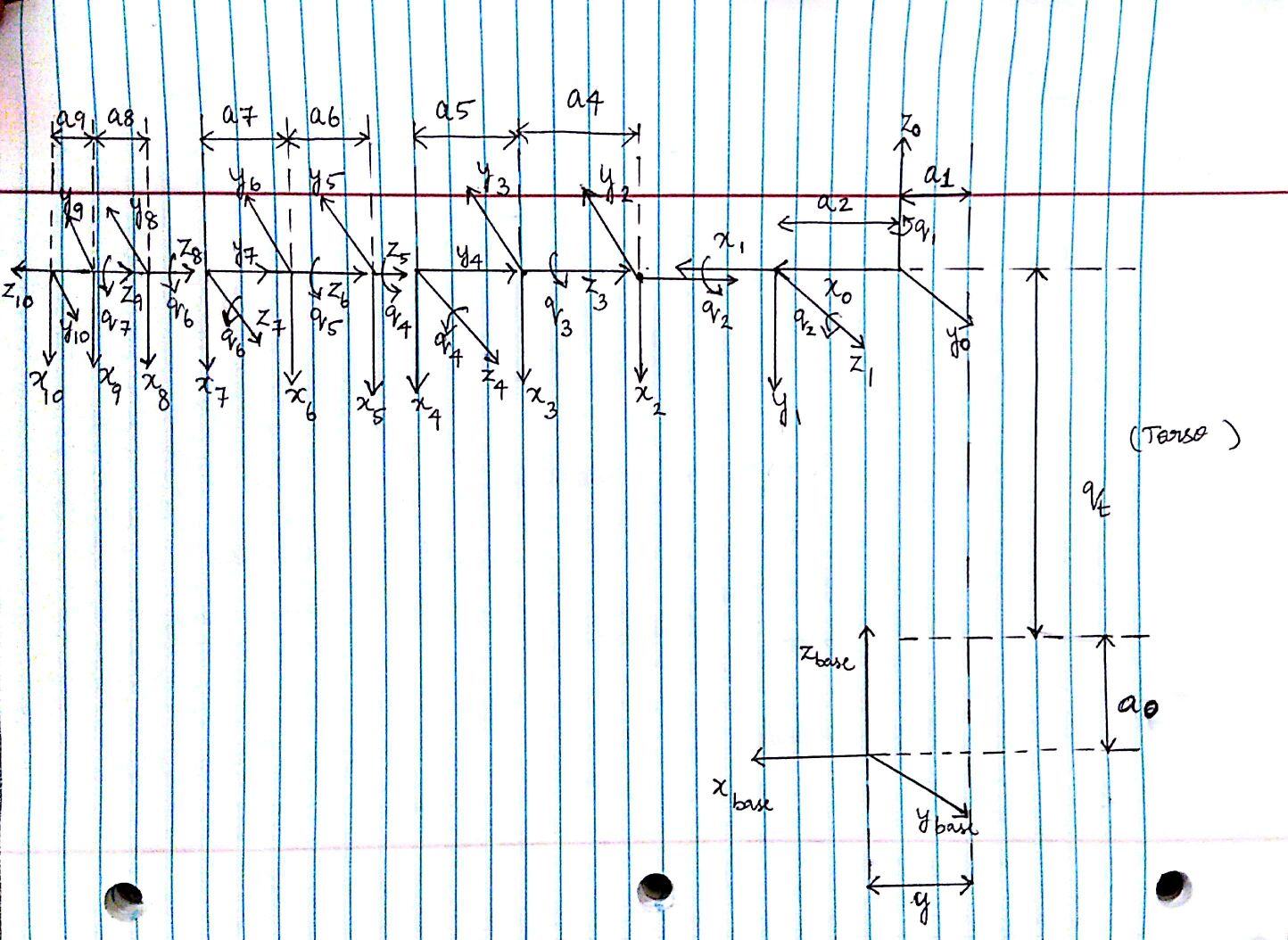}
\caption{Robot Frames - DH convention}
\end{figure}

The next step involves finding the transformation from the base frame to end-effector from the DH parameters. This gives us the transformation matrix. The last column from the transformation matrix gives us the forward kinematics of the system. Once we obtain the transformation matrix we can differentiate it with respect to each of the each of the joint angles to obtain the Jacobian matrix which relates the end-effector velocity with the joint velocities. We can then compute the pseudoinverse of the Jacobian matrix to obtain the joint velocities if we know the end-effector velocities. The equation relating the joint velocities and the end-effector velocities is equation 1,

\begin{equation}
\begin{bmatrix}
\dot{x}\\
\dot{y}\\
\dot{z}\\
\dot{\omega_{x}}\\
\dot{\omega_{y}}\\
\dot{\omega_{z}}\\
\end{bmatrix}= 
J *
\begin{bmatrix}
\dot{\theta_{0}}\\
\dot{\theta_{1}}\\
\dot{\theta_{2}}\\
\dot{\theta_{3}}\\
\dot{\theta_{4}}\\
\dot{\theta_{5}}\\
\dot{\theta_{6}}\\
\end{bmatrix}
\end{equation}

where,
J is the velocity Jacobian and $\dot{\theta}$ represents the joint velocities and the left hand side is the end-effector velocities.

Now we can compute the Kinetic and Potential energies of the system from the joint velocities obtained and the parameters of the Fetch robot provided in their documentation. We can compute the Lagrangian as L = K - P, where K is the Kinetic Energy and P is the potential energy. The torque can be computed from the Lagrangian as shown in equation 2,

\begin{equation}
\tau = \dv{}{t}\pdv{L}{\dot{\theta}} - \pdv{L}{\theta}
\end{equation}

From the torque equation, we can compute the Inertia (M), Coriolis (C) and Gravity (G) matrices and the state space form of the system is written. The torque represented in the form of Inertia, Coriolis and Gravity matrix is shown in equation 3,

\begin{equation}
M(q)\ddot{q} + C(q, \dot{q})\dot{q} + g(q) = \tau
\end{equation}

Figure 4 shows the joints on the Fetch robot using which we obtain the transformation from the robot base to the robot end-effector. 

\begin{figure}[h]
\centering
\includegraphics[width=0.45\textwidth]{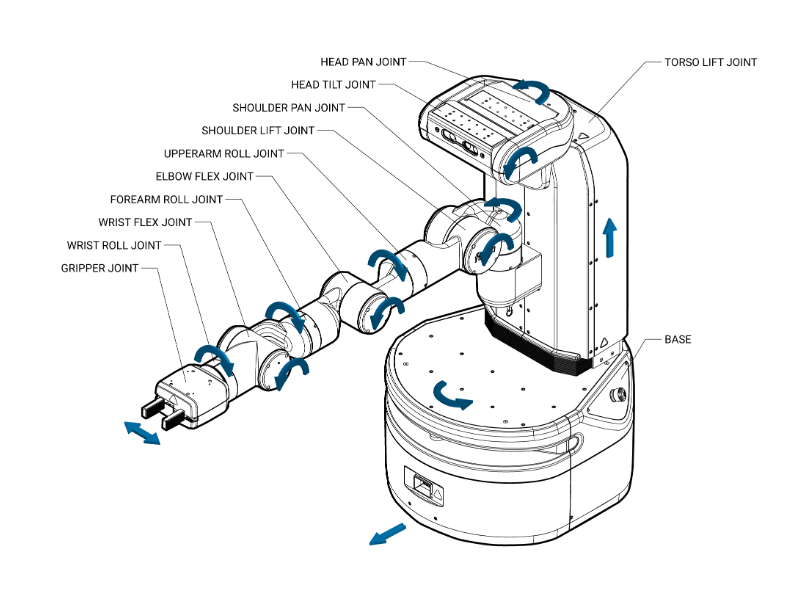}
\caption{Joint Diagram of the Fetch Robot}
\end{figure}

\subsection{Trajectory Planning of manipulator}
We perform point to point motion for each of the seven joints of the manipulator. In our case, we have constraints on position, velocity and acceleration due to the fact that discontinuities in acceleration can lead to impulsive jerk ( derivative of acceleration), which may excite vibrational modes in the manipulator and reduce tracking accuracy. Therefore, we add constraints on the acceleration as well as on the position and velocity thus giving us a quintic polynomial trajectory planning. So, in our case, we have six constraints (one each for initial and final configurations, initial and final velocities, and initial and final accelerations). Therefore we require a fifth order polynomial as given in the equation below,

\begin{equation}
    q(t) = a_{0} + a_{1}t + a_{2}t^2 + a_{3}t^3 + a_{4}t^4 + a_{5}t^5
\end{equation}

where,
$\tau$ is the time of execution of the trajectory and $a_{0}, a_{1}, a_{2}, a_{3}, a_{4}, a_{5}$ are the coefficients of the quintic polynomial.

The problem here is to find a trajectory that connects an initial to a final configuration while satisfying other specified constraints at the endpoints (e.g., velocity and/or acceleration constraints). Without loss of generality, we will plan the trajectory for a single joint, since the trajectories for the remaining joints will be created independently and in exactly the same way. Thus, we will concern ourselves with the problem of determining q(t), where q(t) is a scalar joint variable. The equations for the initial and final position, velocities and accelerations are given in equations 5, 6, 7, 8, 9 and 10. The velocities and the accelerations are obtained by differentiating the joint positions once and twice respectively with respect to time.

\begin{equation}
    q_{0} = a_{0} + a_{1}t_{0} + a_{2}t_{0}^2 + a_{3}t_{0}^3 + a_{4}t_{0}^4 + a_{5}t_{0}^5
    \end{equation}
\begin{equation}
    v_{0} = a_{1} + 2a_{2}t_{0} + 3a_{3}t_{0}^2 + 4a_{4}t_{0}^3 + 5a_{5}t_{0}^4
\end{equation}
\begin{equation}
    \alpha_{0} = 2a_{2} + 6a_{3}t_{0} + 12a_{4}t_{0}^2 + 20a_{5}t_{0}^3
\end{equation}
\begin{equation}
    q_{f} = a_{0} + a_{1}t_{f} + a_{2}t_{f}^2 + a_{3}t_{f}^3 + a_{4}t_{f}^4 + a_{5}t_{f}^5
    \end{equation}
\begin{equation}
    v_{f} = a_{1} + 2a_{2}t_{f} + 3a_{3}t_{f}^2 + 4a_{4}t_{f}^3 + 5a_{5}t_{f}^4
\end{equation}
\begin{equation}
    \alpha_{f} = 2a_{2} + 6a_{3}t_{f} + 12a_{4}t_{f}^2 + 20a_{5}t_{f}^3
\end{equation}

This can be represented in matrix form as follows and therefore if we know the time of planning for each joint, we are able to find the the coefficients of the polynomial which can then be substituted in the above equations and the position, velocity and acceleration for each joint can be figured out.
\begin{equation}
\begin{bmatrix}
q_{0}\\
v_{0}\\
\alpha_{0}\\
q_{f}\\
v_{f}\\
\alpha_{f}\\
\end{bmatrix}= 
\begin{bmatrix}

1 &t_{0} &t_{0}^2 &t_{0}^3 &t_{0}^4 &t_{0}^5\\
0 &1 &t_{0} &3t_{0}^2 &4t_{0}^3 &5t_{0}^4\\
0 &0 &1 &6t_{0} &12t_{0}^2 &20t_{0}^3\\
1 &t_{f} &t_{f}^2 &t_{f}^3 &t_{f}^4 &t_{f}^5\\
0 &1 &t_{f} &3t_{f}^2 &4t_{f}^3 &5t_{f}^4\\
0 &0 &1 &6t_{f} &12t_{f}^2 &20t_{f}^3\\
\end{bmatrix}
\begin{bmatrix}
a_{0}\\
a_{1}\\
a_{2}\\
a_{3}\\
a_{4}\\
a_{5}\\
\end{bmatrix}
\end{equation}

\begin{figure}[h]
\centering
\includegraphics[width=0.45\textwidth]{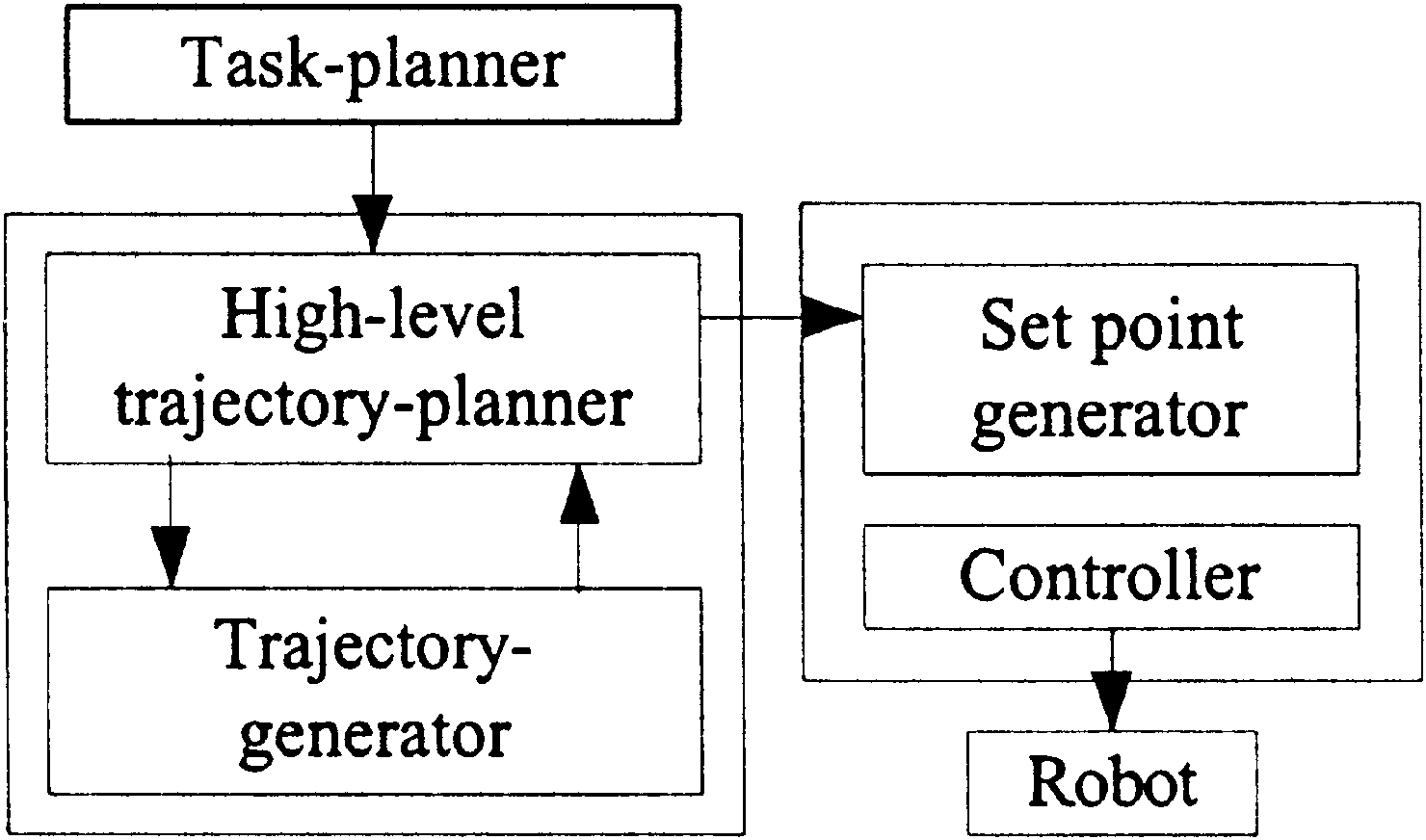}
\caption{Block Diagram of trajectory planner}
\end{figure}

Figure 5 shows the high level control used for the case of trajectory planning and control of the manipulator. We get the high level trajectory plan from the quintic polynomial trajectory for each joint and then a PID controller for each joint make the joint follow the trajectory generated by the quintic polynomial. The PID control law is given by equation 12. This torque is given as feedback to the control law as defined in equation 3.

\begin{equation}
\tau = K_{p}\tilde{q} + K_{v}\dot{\tilde{q}} + K_{i}\int_{0}^{t} \tilde{q}(\sigma) d\sigma
\end{equation}

where,
$K_{p}, K_{v}$ and $K_{i}$ are the position, velocity and integral gains respectively and $\tilde{q}$ is the error term.

\subsection{Trajectory Planning of Mobile Base}
The trajectory planning for the mobile base is performed with the A* algorithm. A* is a search algorithm that is widely used in path finding and graph traversal. Peter Hart, Nils Nilsson and Bertram Raphael first described the algorithm in 1968. As A* traverses the graph, it follows a path of the lowest known cost, keeping a sorted priority queue of alternate path segment along  the  way.  If,  at  any  point,  a  segment of  the  path  being  traversed  has  a  higher  cost than another encountered path segment, it abandons the higher-cost path segment and traverses the lower-cost path segment instead.  This process continues until the goal is reached [16]. It utilizes  a  heuristic  to  focus  the  search  towards  the  most promising areas of the search space. A* aims to find an optimal path which may not be feasible given time constraints and the dimensionality of the problem. The priority queue is based on f = g + h, where g is the backward cost and h is the forward cost to goal or called the heuristic. For the A* algorithm to give an optimal solution in the case of the tree search, the heuristic should be admissible (i.e. it should never overestimate the true cost to the goal). It is the user who has to provide the heuristic which is admissible.

The path returned by the A* planner is then time-parametrized before being passed onto the controllers which takes the path returned from the planner and the actual path followed by the robot. The error term for the controller in this case is the euclidean distance between the time-parametrized path from the A* planner and the actual path followed by the mobile base. 

\subsection{Inverse Kinematics}
We make use of the Trac\_ik package available in the ROS framework to perform Inverse Kinematics of the manipulator arm.  Trac\_ik is an enhancement of the KDL's (Kinematics and Dynamics Library) pseudoinverse Jacobian solver which improves the performance drastically. Some of the issues with the KDL slover which are addressed, the KDL Inverse Kinematics takes as input a maximum number of iterations to try which mainly depends on the time to compute the Jacobian. Thus, there is no need to give the user the criteria to choose the number of iterations which can be decided by our solver where the user is only prompted to provide the maximum time allowed for solving. This is functionally equivalent to running KDL with a number of iterations, but the maximum number of iterations is dynamically determined based on the user-desired solve time.

The Inverse Jacobian IK implementations can get stuck in local minima during the iteration process. This is valid in cases where joints have physical limits on their range. In the KDL implementation, there is nothing to detect local minima or mitigate this scenario. In this solver, the local minima are detected and mitigated by changing the next seed. Consequently, the performance of the KDL IK solver can be significantly improved by simply using random seeds for q to improve the performance of the iterative algorithm when local minima are detected. But this method does not handle the constraints of joint limits very well.

One way to avoid these failures is to solve IK using methods that better handle constraints like joint limits. IK as a nonlinear optimization problem can be solved locally using sequential quadratic programming (SQP). SQP is an iterative algorithm for nonlinear optimization, which can be applied to problems with objective functions and constraints that are twice continuously differentiable. SQP minimizes the overall amount of joint movement and only considers Cartesian error as a constraint. The equation for the SQP problem in our case is given in equation 13,

\begin{equation}
\begin{split}    
   \argmin_{q \in R^n} \quad (q_{seed} - q)^T\; (q_{seed} - q) \\ s.t. \quad f_{i}(q) \leq b_{i}, i = 1, ..., m
   \end{split}
\end{equation}

where $q_{seed}$ is the n-dimensional seed value of the joints, and the inequality constraints $f_{i}(q)$ are the joint limits, the Euclidean distance error, and the angular distance error.

SQP method uses dual quaternion error as a combined measure of distance error and angular error in Cartesian space. Generating the dual quaternion error takes non-trivial computational time compared to simpler metrics such as the sum of squares metric. The SQP-SS (Sum of Squares)used the sum of squares metric when the sum of the squares of the error are considered. The Sum of Squares equation is given in equation 14,

\begin{equation}
\phi_{SS} = p_{err} * p_{err}^T
\end{equation}
    
Therefore, for a single IK solver request Trac\_ik spawns two solvers, one running SQP-SS and the other running KDL. Once either finishes with a solution, both threads are immediately stopped, and the resulting solution is returned. This is the same approach that we are following in our project and the results for the IK solver are discussed in a later section.

\subsection{State Machine}

A finite state machine (FSM) is a mathematical model of computation. It is an abstract machine that can be in exactly one of a finite number of states at any given time. The FSM can change from one state to another in response to some external inputs. the change from one state to another is called a transition. A FSM is defined by a list of its states, its initial state, and the conditions for each transition. We have defined a state machine for the pick and place task using the Fetch robot and it can be defined in Figure 6. First, we check the robot status which gives us the information of the location of the base and the location of the end-effector. Various states call different parts of our system, thus making the execution of the overall task possible. The transition for every state is provided by a feedback mechanism as defined in the ROS framework.  

\begin{figure}[h]
\centering
\includegraphics[width=0.45\textwidth]{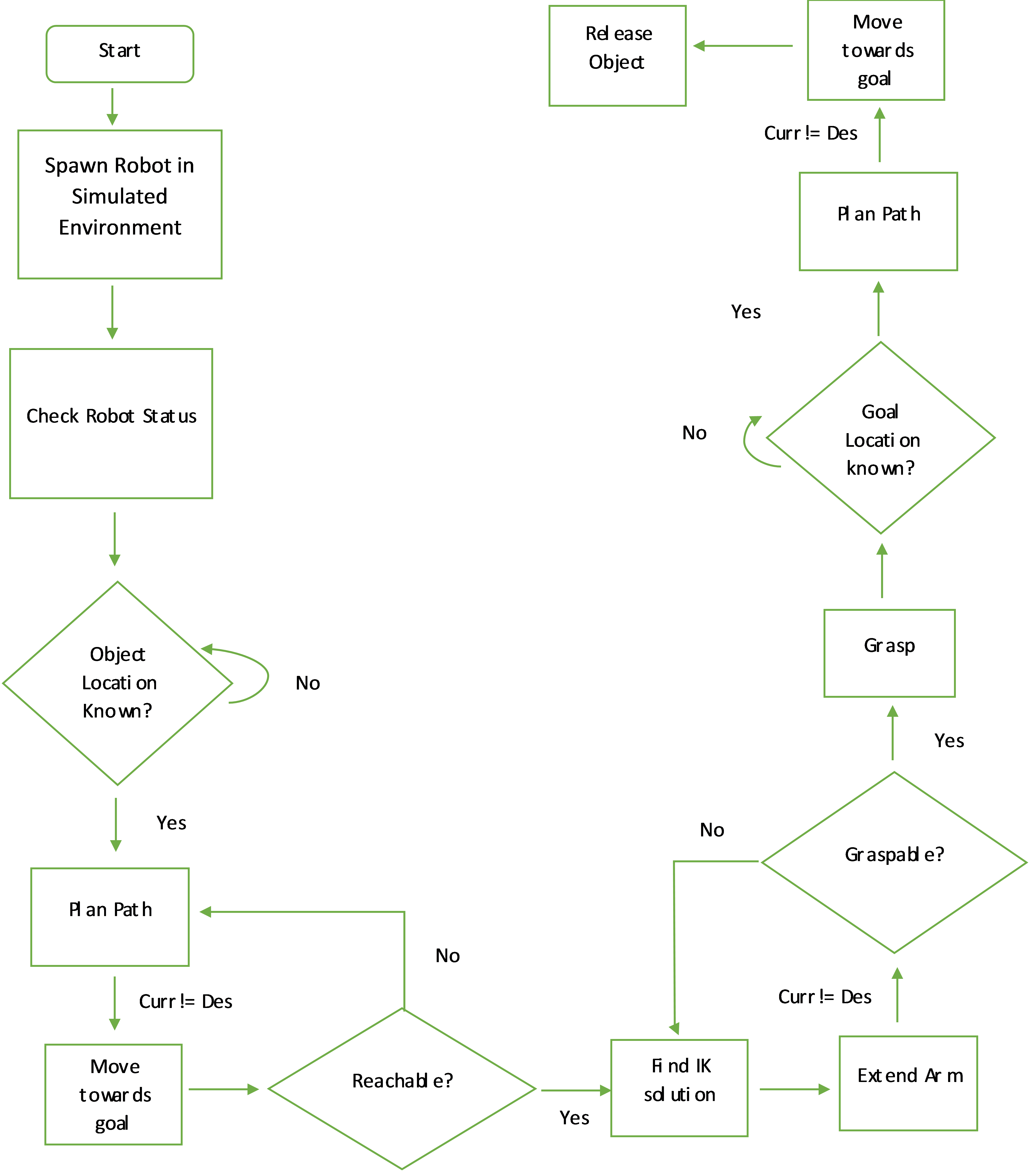}
\caption{State Machine for the pick and place task}
\end{figure}

\subsection{Implementation Platform}

The platform that we use to test our system is the Fetch robot. The Fetch kinematics are defined by the robot URDF (unified robot description format) model which specifies the attributes of the joints, links, and frames of the robot. A link element in the URDF describes a rigid body with inertia, visual features, and coordinate frames. A joint element in the URDF defines the kinematics, dynamics, safety limits, and type. We entered these information based on the DH parameters and the system dynamics derived earlier. Then we define the direction of the coordinate frames for the Fetch robot.

The coordinate frames for all links in the Fetch and Freight are defined with positive z-axis up, positive x-axis forward, and positive y-axis to the robot-left when Fetch is in the home pose. Joint position and velocity limits are enforced for all the joints in the Fetch robot. This acts the constraints during trajectory planning and finding inverse kinematic solutions. We make use of various sensors available on the Fetch robot to perform the trajectory control. The sensors which are useful in our project are described below.

The Fetch robot has the following sensors which is of use in our system, a SICK TIM571 scanning range finder, a 6-axis inertial measurement unit (IMU), Primesense short-range RGBD sensor and a Gripper sensor. The laser has a range of 25m, 220° field of view, 15Hz update rate and angular resolution of 1/3°. The laser publishes both distance and RSSI (Received Signal Strength Indication) to the base\_scan topic,  The gyroscope within the IMU is capable of measuring +/-2000 degrees per second, while the accelerometers are capable of measuring +/-2g.  In addition to the position and effort feedback of the gripper joint, the gripper incorporates a 6-axis inertial measurement unit (IMU).

We performed the entire task in ROS/Gazebo with the Fetch robot in a simulation environment. The results obtained after performing our experiments are explained in Section IV.

\section{Results}
In this project, the first task of deriving the system dynamics began with formulating the DH parameters for the manipulator of fetch robot as discussed in the methodology. The DH parameters derived for each of the consecutive links are shown in figure 7. In figure 7 'D' stand for dummy frame. The Derived DH parameters were validated by performing forward kinematics in MATLAB.
\begin{figure}
\includegraphics[width=0.45\textwidth]{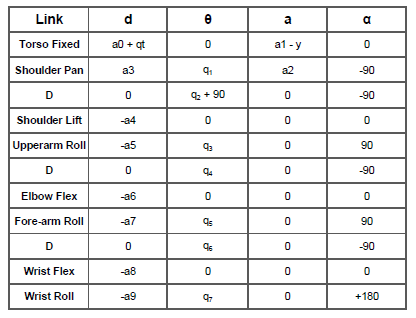}
\caption{DH Parameters}
\end{figure}

 The task of creating the simulation world with Fetch robot and the environment was done in gazebo. The world environment was formed by placing models of the shelf and object in gazebo world. We created a world with the shelf and the object and also spawned the fetch robot to this world by importing it’s URDF file in the simulation environment. Figure 3 \& 4 shows screen shot of the gazebo world with the Fetch robot, Shelf and the object.

\begin{figure}
\begin{minipage}[c]{0.22\textwidth}
    \includegraphics[width=\textwidth]{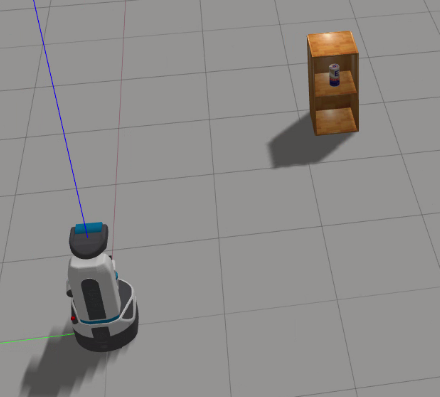}
  \end{minipage}
 \hspace{0.05\linewidth}
  \begin{minipage}[c]{0.22\textwidth}
    \includegraphics[width=\textwidth]{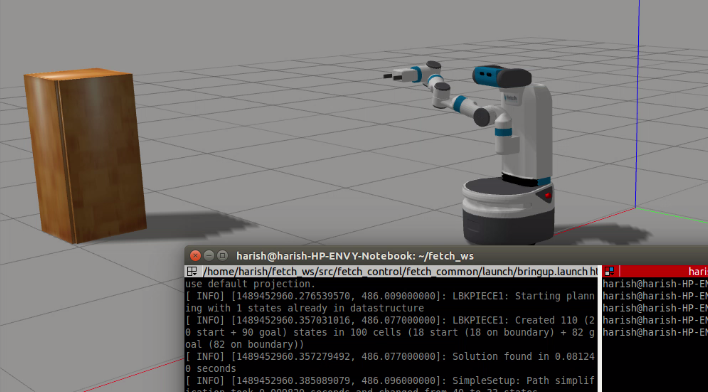}
  \end{minipage}
\caption {Simulation World}
\end{figure}
\begin{figure}
\includegraphics[width=0.45\textwidth]{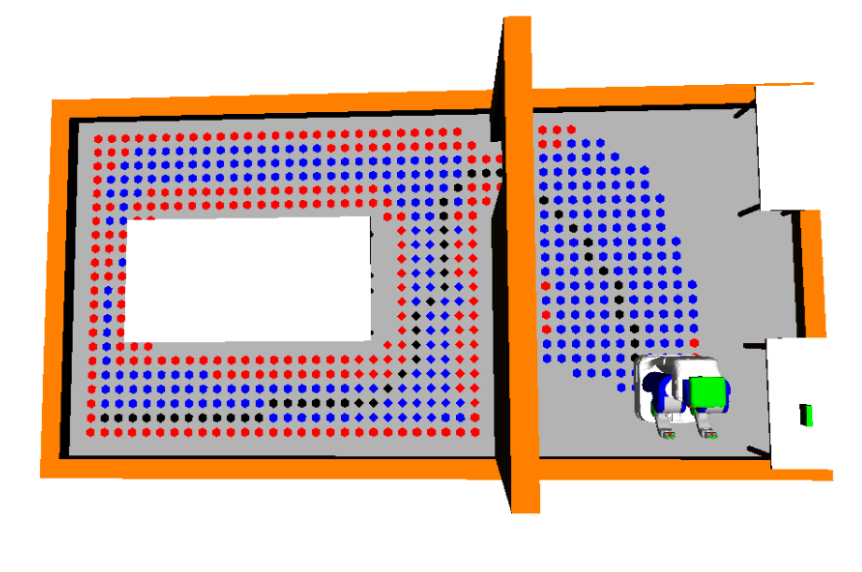}
\caption{Visualization of the path found by A* for an 8-connected Euclidean heuristic is shown in black}
\end{figure}
After this step, when the base of the robot moves we do inverse kinematics with the method explained in the previous section. We used the open source package Trac\_ik solver in the ROS framework. Trac\_ik performed very well compared to other classical solvers such as Orocos KDL solver etc in our case. Trac\_ik took an average time of 0.44ms to find an Inverse Kinematic solution when run for 100 trials against the Orocos KDL solver which took 0.72 ms for the same 100 trials. The success rate of the Trac\_ik solver was also higher when compared to the Orocos KDL solver.

Next, we did trajectory planning for both the manipulator in the fetch robot as well as the mobile base. Trajectory planning for the manipulator was done in joint space considering the joint limit constraints.This trajectory planning was done in joint space so that there is no need to perform inverse kinematics operation at each instant of planning which can be costly. This saves considerable amount of time during planning. Moreover, inverse kinematics can give multiple solutions and obtaining the correct set of joint angles will be time consuming. Planning in joint space rather than Cartesian space saved us time. The position, velocity and acceleration profiles obtained for one of the joints is shown in figures 10 and 11. Similar profiles were obtained for all the joints. We performed quintic polynomial planning because when we don't consider the acceleration of the joint, it leads to a lot of jerkiness in the system. The next part was the trajectory planning for the mobile base of the Fetch robot.

\begin{figure}[h]
\centering
\includegraphics[width=0.4\textwidth]{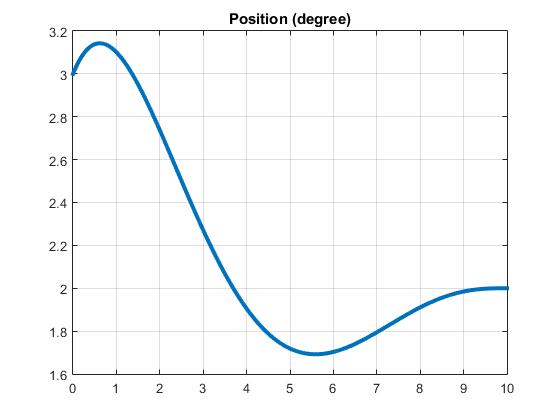}
\caption{Position for each joint after Trajectory planning}
\end{figure}
\begin{figure}

  \begin{minipage}[c]{0.22\textwidth}
    \includegraphics[width=\textwidth]{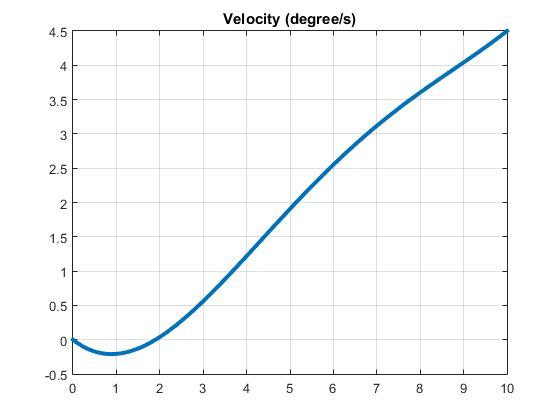}
  \end{minipage}
   \hspace{0.05\linewidth}
    \begin{minipage}[c]{0.22\textwidth}
    \includegraphics[width=\textwidth]{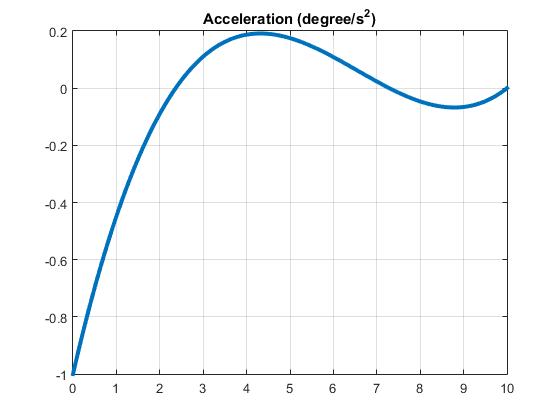}
  \end{minipage}
\caption {Velocity and Acceleration profile for each joint after Trajectory Planning}
\end{figure}

The trajectory planning for the mobile base was done using A* path planning algorithm as discussed in methodology. In order to visualize the working of the A* algorithm we tested our algorithm first on a PR2 robot in a simulation environment known as OpenRAVE. The A* algorithm was written as plugin in the global planner library available in ROS. The path was generated by the A* planning algorithm for four connected space as well as eight connected space. We used the manhattan and euclidean heuristic for both of these scenarios and tested the performance in OpenRAVE.  In case of eight connected we observed euclidean heuristic to be superior. Number of nodes explored in case of Manhattan heuristic was 3462. Number of nodes explored in case of  Euclidean heuristic was 4555. Cost in case of Manhattan was 126.70 and the cost in case of Euclidean was 104.83 for the eight connected space. Therefore, we used eight connected Euclidean heuristic and performed the planning. The path generated by A* planning algorithm for an eight connected space with euclidean distance heuristic is shown in figure 9.

Now, in order to move the base of the fetch robot, we published the parameterized path found by the A* algorithm on a topic to which the base subscribed. Before doing this, we found the transformation between the base frame and the odometry frame. This was found using the tf package available in ROS where our tf class had a subscriber which waits for the transform between the odometry and the base frame and records it. Then the geometry\_msgs of type Twist was published on the cmd\_vel topic to move the robot base. This method was used to move the robot base to the desired location according to the 

The PID controller is implemeted for both the trajectory planning of the manipulator and the mobile base. After getting the trajectory for each joint from the planner, we use a PID controller to ensure that each joint follows the trajectory generated by the planners. In order to make the system stable we tune the PID controller by modifying the controller gains. The trajectory which is generated for the mobile base is time parameterized and given to the PID controller as input. The PID controller minimizes the  euclidean distance  between  the time-parameterized  path  from  the  A* planner and the actual path followed by the mobile base. The joint trajectories, Velocity profile and joint efforts of Shoulder Pan joint, Shoulder Lift joint, Upperarm Roll joint, Elbow Flex joint and Forearm Roll joint are shown in figures 12, 13 and 14 respectively.

\begin{figure}[h]
\centering
\includegraphics[width=0.4\textwidth]{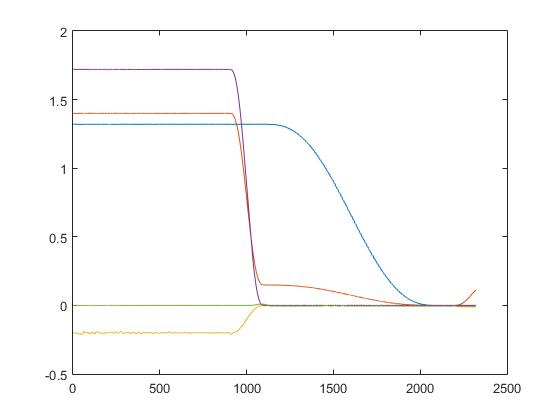}
\caption{Joint Trajectories of each joint}
\end{figure}

\begin{figure}[h]
\centering
\includegraphics[width=0.4\textwidth]{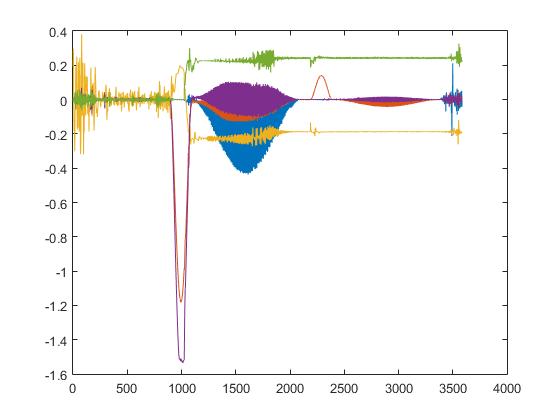}
\caption{Velocity Profile of each joint}
\end{figure}

\begin{figure}[h]
\centering
\includegraphics[width=0.4\textwidth]{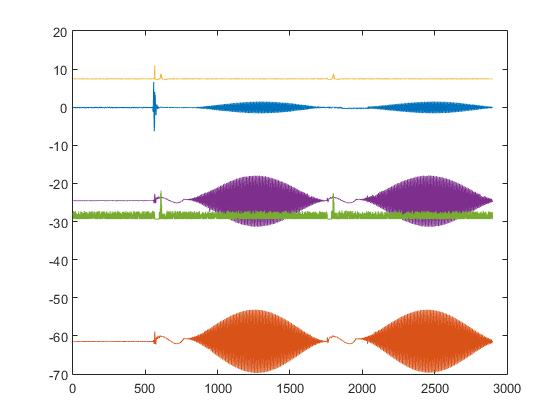}
\caption{Joint efforts for each joint}
\end{figure}

\begin{figure}
  \includegraphics[width=0.45\textwidth]{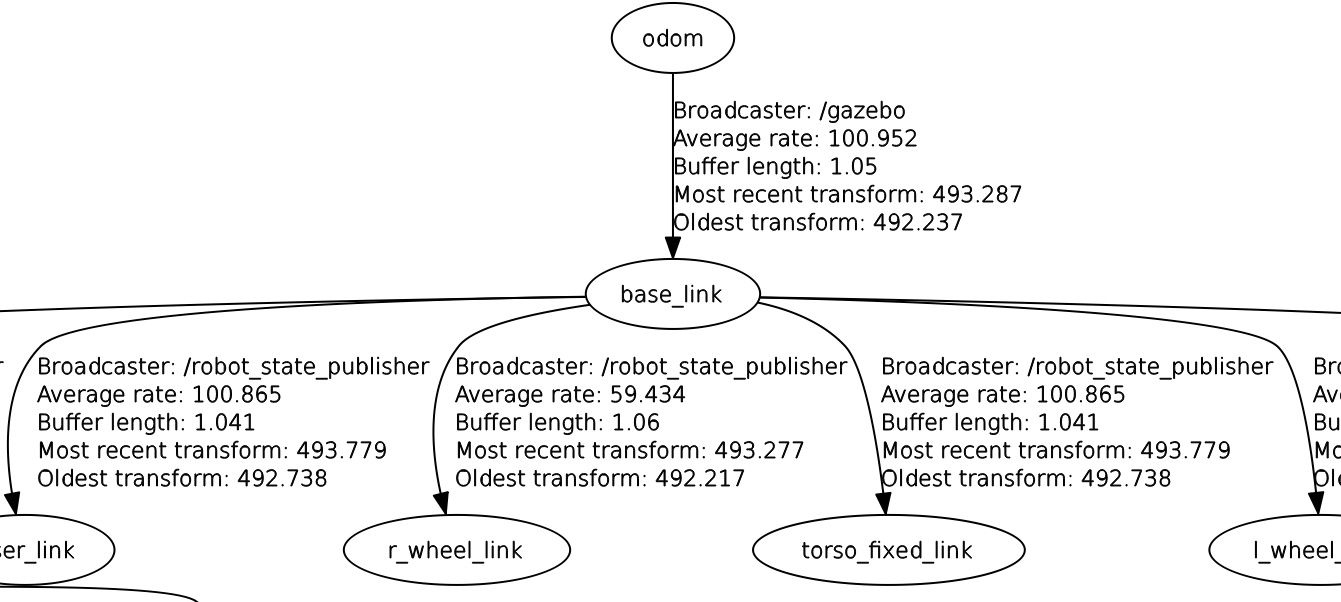}
  \caption{tf Tree}
\end{figure}

\begin{figure}
  \includegraphics[width=0.45\textwidth]{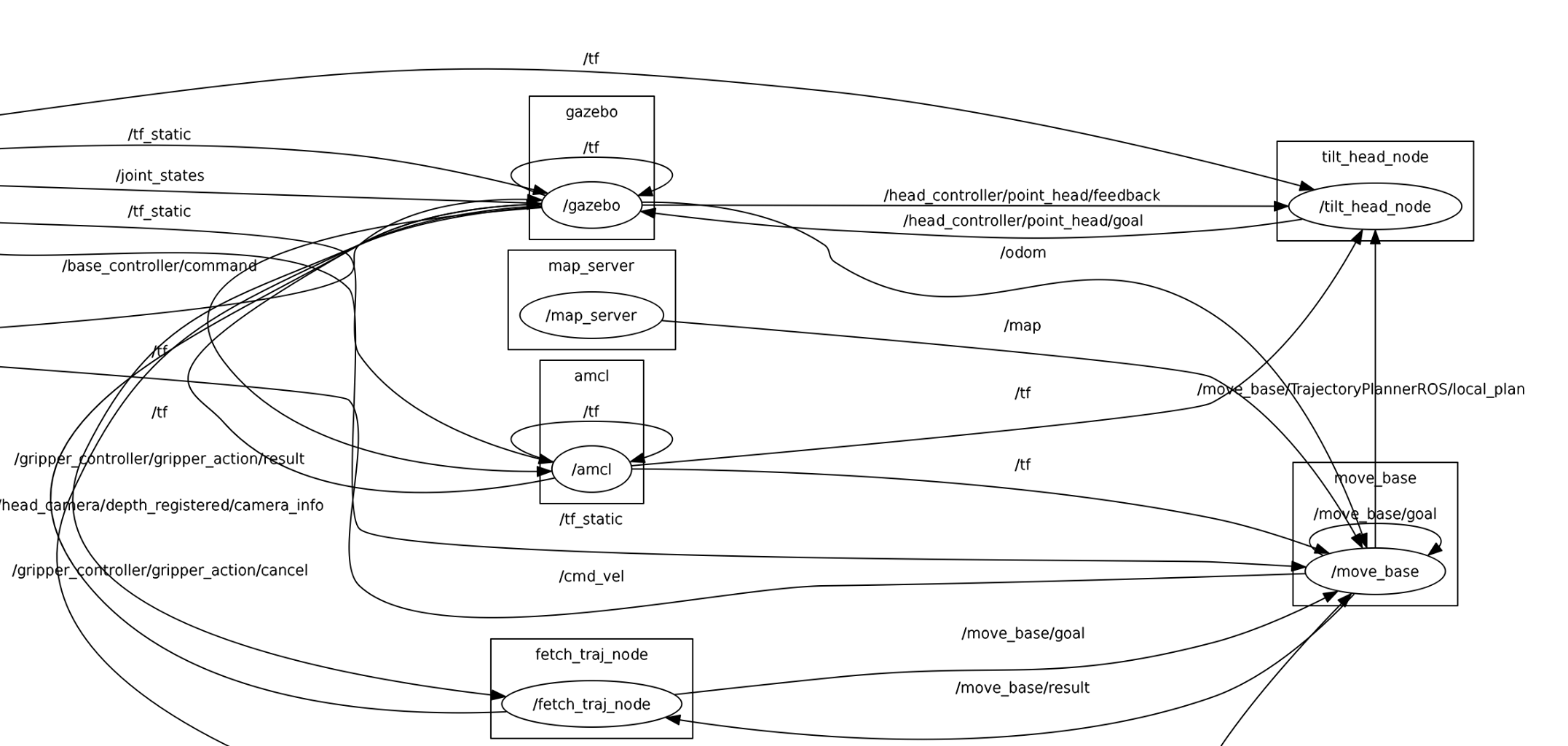}
    \caption{rqt-graph}
\end{figure}

After the planning step, we implemented a state machine in C++ for the pick and place task on the fetch robot. Different states are defined in the state machine and it changes from one state to another when it gets an transition request in the form of a ROS message after the previous state has been completed. The State Machine was used to automate the entire process of the pick and place task. All the tasks which was mentioned above is included in a ROS node named as fetch\_traj\_node. When wrote this finite state machine as a ROS node using the roscpp client library and when we run this node, all the tasks such as trajectory planning of manipulator, trajectory planning of mobile base, inverse kinematics etc. are executed sequentially according to the flowchart explained earlier.

Final task involved controlling the gripping action i.e like controlling the amount by which the gripper closes to grab an object on the shelf. We used the gripper\_controller or gripper\_action interfaces to send gripper commands using the control\_msgs topic. The gripper command takes in position and effort as parameters. Generally, the gripper is commanded to a fully closed or fully opened position, so effort is used to limit the maximum effort. As the gripper never fully reaches the closed position, the grasp strength will be determined by the maximum effort. So, in order to pick up objects we need to control the effort generated by the gripper in the fetch robot.

The tf tree and rqt graph showing the transformation from one frame to another as well as the list of active nodes and their subscribers and publisher topic are shown in figures 15 and 16 respectively.

\section{Conclusion and Future Work}
In this project, we have implemented Trajectory Control for the Fetch robot which is a differential drive mobile manipulator robot in a simulated environment. We have shown results and implementation of the dynamics of the fetch robot, Trajectory planning for the manipulator arm using a quintic spline and mobile base trajectory planning has been implemented with the A* planning algorithm and the results have been shown in Section IV. The state machine for the task of indoor warehouse object manipulator has been carried out and results of it's implementation has also been discussed. Also, the Trac\_ik package has been used for performing the inverse kinematic calculations for the manipulator of the robot. Overall, the project goals stated in Section I have been implemented and the corresponding methodology shown in Section III and the results of our experiments carried out discussed in Section IV.

In the future, we would develop a global planner which would just get the location of an object in an environment and then autonomously determine the amount at which the base should be moved and then the end-effector. Right now, we don't have this feature as we plan for the mobile base and the manipulator separately. Also, we would like to make our trajectory control package open source since there is no robust package that achieves this efficiently. But, more work is needed to make it generic for all robots. We would implement our system on an actual hardware platform in an indoor warehouse environment to test it's performance. We have preformed the picking of the object but we would want to incorporate computer vision techniques to make the state machine more powerful. One more extension of our work would be to make all the different aspects modular so that it can be ported to any platform easily. 

\section{Acknowledgements}
We would like to thank Professor Eugene Eberbach of the department of Robotics Engineering at Worcester Polytechnic Institute for his patience, guidance, advice and support over the term of the project, and Junius Santoso for providing us useful inputs facilitating the successful completion of our project.


\begin{thebibliography}{1}

\bibitem{}
 Correll, N., Bekris, K.E., Berenson, D., Brock, O., Causo, A., Hauser, K., Okada, K., Rodriguez, A., Romano, J.M. and Wurman, P.R., 2016. Lessons from the amazon picking challenge. arXiv preprint arXiv:1601.05484
\bibitem{}
 Papadopoulos, E. and Poulakakis, J., 2000, July. Trajectory planning and control for mobile manipulator systems. In The 8th IEEE Mediterranean Conf. on Control \& Automation.
\bibitem{}
  Tchon, Krzysztof and Janusz Jakubiak. “An Extended Jacobian Inverse Kinematics Algorithm for Doubly Nonholonomic Mobile Manipulators.” ICRA (2005).
\bibitem{}
  Mazur, A. and Arent, K., 2006. Trajectory tracking control for nonholonomic mobile manipulators. In Robot Motion and Control (pp. 55-71). Springer London.
\bibitem{}
Solea, R. and Cernega, D., 2011. Trajectory Tracking Control of Mobile Manipulators based on Kinematics. In ICINCO (2) (pp. 21-27).
\bibitem{}
Ide, S., Takubo, T., Ohara, K., Mae, Y. and Arai, T., 2011, November.
Real-time trajectory planning for mobile manipulator using model
predictive control with constraints. In Ubiquitous Robots and Ambient
Intelligence (URAI), 2011 8th International Conference on (pp. 244-249).
IEEE.
\bibitem{}
Dong, W., Xu, Y. and Wang, Q., 2000. On tracking control of mobile
manipulators. In Robotics and Automation, 2000. Proceedings. ICRA’00.
IEEE International Conference on (Vol. 4, pp. 3455-3460). IEEE.
\bibitem{}
Korayem, M.H. and Nohooji, H.R., 2008, October. Trajectory optimization
of flexible mobile manipulators using open-loop optimal control
method. In International Conference on Intelligent Robotics and Applications
(pp. 54-63). Springer Berlin Heidelberg.
\bibitem{}
Korayem, M.H., Nazemizadeh, M. and Azimirad, V., 2011. Optimal
trajectory planning of wheeled mobile manipulators in cluttered environments
using potential functions. Scientia Iranica, 18(5), pp.1138-1147.
\bibitem{}
Yu, Q. and Chen, I.M., 2002. A general approach to the dynamics of nonholonomic mobile manipulator systems. TRANSACTIONS-AMERICAN SOCIETY OF MECHANICAL ENGINEERS JOURNAL OF DYNAMIC SYSTEMS MEASUREMENT AND CONTROL, 124(4), pp.512-521.
\bibitem{}
Beeson, P. and Ames, B., 2015, November. TRAC-IK: An open-source library for improved solving of generic inverse kinematics. In Humanoid Robots (Humanoids), 2015 IEEE-RAS 15th International Conference on (pp. 928-935). IEEE.
\bibitem{}
S.R. Buss, Introduction  to  Inverse  Kinematics  with  Jacobian  Transpose, Pseudoinverse  and  Damped  Least  Squares  methods. Survey, University of California, 2004.
\bibitem{}
P. Beeson, https://bitbucket.org/traclabs/tracik.git, accessed: 2016-07-22.
\bibitem{}
ROS Framework, http://www.ros.org, accessed: 2016-07-22.
\bibitem{}
Orocos  Kinematics  and  Dynamics,  http://www.orocos.org,  accessed: 2016-07-22.
\bibitem{}
Nosrati, M., Karimi, R. and Hasanvand, H.A., 2012. Investigation of the*(star) search algorithms: Characteristics, methods and approaches. World Applied Programming, 2(4), pp.251-256.


\end{thebibliography}
\end{document}